\documentclass[10pt,journal]{IEEEtran}

\usepackage{cite}
\usepackage{amsmath,amssymb,amsfonts}
\usepackage{graphicx}
\usepackage{algorithm}
\usepackage{algpseudocode} 
\usepackage{hyperref}
\hypersetup{hidelinks=true}
\usepackage{textcomp}
\usepackage{booktabs}
\usepackage{placeins}
\usepackage{multirow}
\usepackage{enumerate}
\usepackage{soul}

\def\BibTeX{{\rm B\kern-.05em{\sc i\kern-.025em b}\kern-.08em
    T\kern-.1667em\lower.7ex\hbox{E}\kern-.125emX}}
\begin{document}

\title{Federated Learning for Epileptic Seizure Prediction Across Heterogeneous EEG Datasets}
\author{Cem Ata Baykara, Saurav Raj Pandey, Ali Burak Ünal, Harlin Lee, and Mete Akgün
\thanks{The first two student authors and the last two authors contributed equally. This work was partially supported by the UNC-Tübingen Seed Fund for Collaborations in Data Science and by the German Federal Ministry of Education and Research (BMBF) under project number 01ZZ2010 (MDPPML) and 01ZZ2316D (PrivateAIM).}
\thanks{Baykara, Ünal and Akgün are with the Department of Computer Science, Medical Data Privacy and Privacy-Preserving Machine Learning (MDPPML) Group and Institute for Bioinformatics and Medical Informatics (IBMI), University of Tübingen, Sand 14, 72076 Tübingen, Germany (cem.baykara@uni-tuebingen.de, ali-burak.uenal@uni-tuebingen.de, mete.akguen@uni-tuebingen.de).}
\thanks{Pandey and Lee are with the Department of Computer Science and the School of Data Science and Society, University of North Carolina at Chapel Hill, Chapel Hill, NC, 27599 USA (srpandey@cs.unc.edu, harlin@unc.edu).}}

\maketitle

\begin{abstract}
Developing accurate and generalizable epileptic seizure prediction models from electroencephalography (EEG) data across multiple clinical sites is hindered by patient privacy regulations and significant data heterogeneity (non-IID characteristics). Federated Learning (FL) offers a privacy-preserving framework for collaborative training, but standard aggregation methods like Federated Averaging (FedAvg) can be biased by dominant datasets in heterogeneous settings. This paper investigates FL for seizure prediction using a single EEG channel across four diverse public datasets (Siena, CHB-MIT, Helsinki, NCH), representing distinct patient populations (adult, pediatric, neonate) and recording conditions. We implement privacy-preserving global normalization and propose a Random Subset Aggregation strategy, where each client trains on a fixed-size random subset of its data per round, ensuring equal contribution during aggregation. Our results show that locally trained models fail to generalize across sites, and standard weighted FedAvg yields highly skewed performance (e.g., 89.0\% accuracy on CHB-MIT but only 50.8\% on Helsinki and 50.6\% on NCH). In contrast, Random Subset Aggregation significantly improves performance on under-represented clients (accuracy increases to 81.7\% on Helsinki and 68.7\% on NCH) and achieves a superior macro-average accuracy of 77.1\% and pooled accuracy of 80.0\% across all sites, demonstrating a more robust and fair global model. This work highlights the potential of balanced FL approaches for building effective and generalizable seizure prediction systems in realistic, heterogeneous multi-hospital environments while respecting data privacy.
\end{abstract}


\section{Introduction}
\label{sec:introduction}

Epilepsy is a prevalent neurological disorder characterized by recurrent, unpredictable seizures, significantly impacting the quality of life and safety of millions worldwide \cite{WHO_epilepsy_factsheet}. Accurate prediction of seizures minutes to hours before their onset holds immense clinical value, offering opportunities for timely intervention, reduction of seizure-related injuries, and improved patient autonomy~\cite{kuhlmann2018epileptic}. Deep learning models applied to electroencephalography (EEG) data have shown promise in developing automated seizure prediction systems \cite{jemal2024domain,usman2022epileptic}.

\begin{figure}[t]
    \centering
    \includegraphics[width=0.7\linewidth]{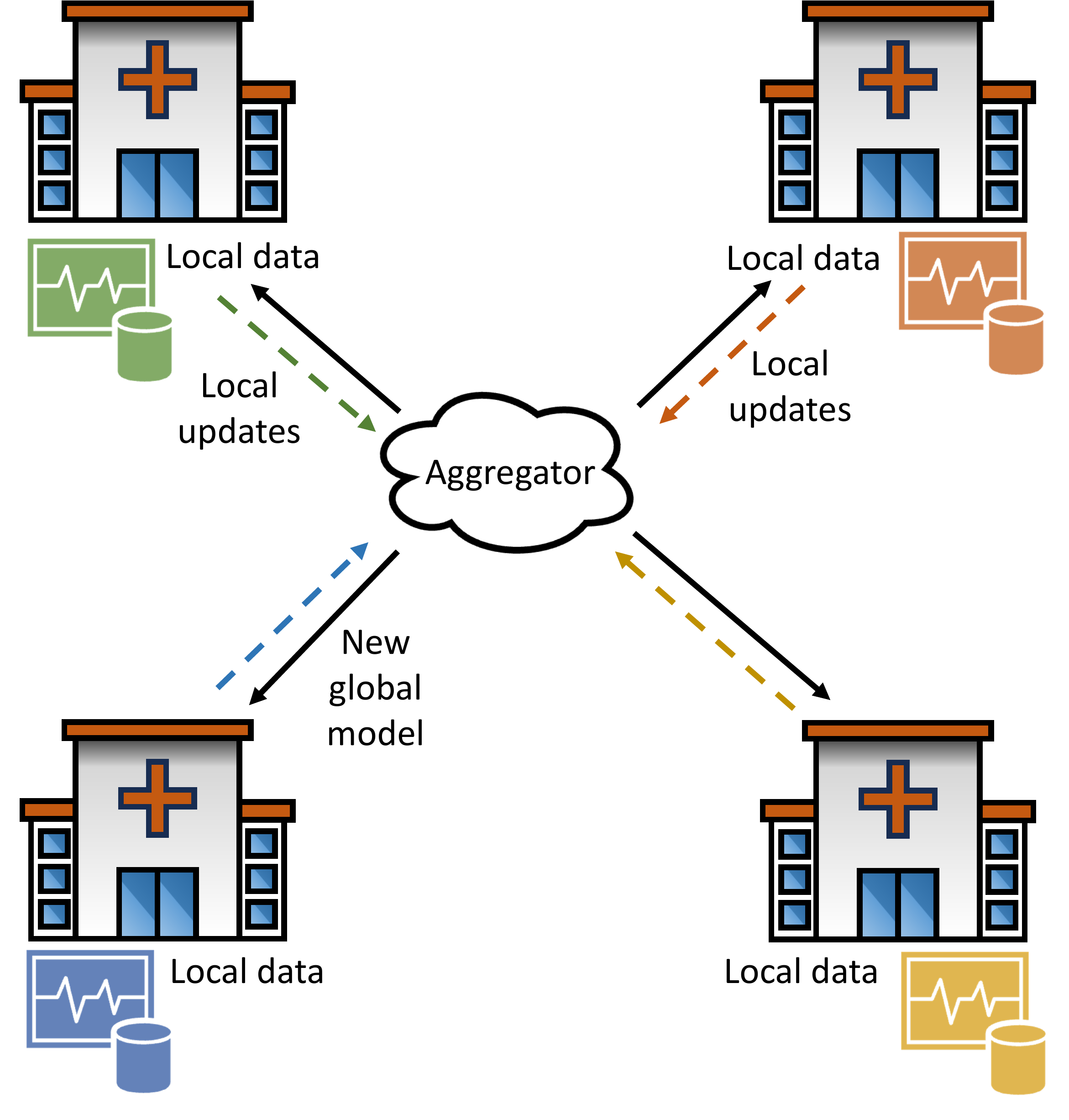} 
    \caption{Illustration of the FL system involving four hospitals with non-IID local datasets and a central aggregator coordinating training. Each hospital sends local model updates to the aggregator and receives the aggregated global model in return.}
    \label{fig:fl_system_architecture}
\end{figure}

However, translating these promising results into widespread clinical practice faces significant challenges. Training robust deep learning models typically requires large, diverse datasets, which are often fragmented across multiple hospitals or clinical centers. Centralizing sensitive patient EEG data for model training raises substantial privacy and security concerns, often conflicting with regulations like HIPAA \cite{HIPAA} and GDPR \cite{GDPR}. Furthermore, even if centralization were possible, models trained on data pooled from various sources may struggle with generalization due to inherent distribution shifts between sites. These shifts arise from differences in patient demographics (e.g., age groups from neonates to adults), EEG acquisition protocols, hardware specifications, and even clinical focus (e.g., sleep studies vs. routine monitoring), leading to models that perform well on average but poorly on specific under-represented patient groups or clinical settings \cite{aminifar2024privacy}. As demonstrated in our preliminary experiments (Section \ref{subsec:localized_learning}), models trained exclusively on data from one hospital often fail to generalize effectively to data from other hospitals.

Federated Learning (FL) \cite{mcmahan2017communication, rieke2020future} addresses privacy concerns by allowing collaborative model training without sharing raw data; instead, local model updates are centrally aggregated. However, data heterogeneity (non-IID data) across participants is a key challenge \cite{kairouz2021advances, li2020federated}, especially in multi-hospital settings with varying patient populations and equipment. Federated Averaging (FedAvg) \cite{mcmahan2017communication}, the standard aggregation method in FL, weighs each client's contribution by its local dataset size. While this improves convergence in IID settings, it can bias the global model toward dominant clients in heterogeneous non-IID scenarios, impairing performance on under-represented sites \cite{zhu2021federated}.

This paper investigates the application of FL for epileptic seizure prediction in a realistic, highly heterogeneous multi-hospital setting. We utilize four publicly available EEG datasets: CHB-MIT (pediatrics) \cite{guttag2010chb}, \cite{shoeb2009application}, Helsinki (neonates) \cite{stevenson2019dataset}, Siena (adults) \cite{detti2020siena}, \cite{siennasecondcite}, and NCH (pediatrics, primarily sleep studies) \cite{lee2022large}, \cite{lee2021nch}, simulating four distinct clinical sites with diverse patient populations and data characteristics \cite{goldberger2000physiobank}. An overview of the federated learning setup is shown in Fig.~\ref{fig:fl_system_architecture}. To enhance practical applicability, we focus on using a single, common EEG channel (F3-C3) and adapt a lightweight deep learning model, TinySleepNet \cite{TinySleepNet}, for the prediction task. Recognizing the challenge posed by non-IID data and varying dataset sizes, we implement a privacy-preserving global normalization technique and propose a novel aggregation strategy called Random Subset Aggregation. This approach aims to mitigate bias by ensuring each hospital contributes equally during the aggregation process, regardless of its local dataset size, thereby yielding a more generalizable and robust global model.

The main contributions of this paper are:
\begin{itemize}
\item The first application, to our knowledge, of FL for seizure prediction across these four specific and highly diverse public EEG datasets (Siena, CHB-MIT, Helsinki, NCH), simulating a realistic cross-hospital collaboration.
\item A clear demonstration of the limitations of purely localized models (poor cross-site generalization) and standard FedAvg (bias towards dominant datasets) in this highly heterogeneous (non-IID) setting.
\item The proposal and evaluation of a Random Subset Aggregation strategy within the FL framework, showing its effectiveness in improving fairness and achieving robust macro-average performance across diverse clients compared to baseline FL methods.
\item Implementation details including the use of a practical single-channel EEG setup (F3-C3), a lightweight model architecture suitable for potential deployment, and a privacy-preserving global data normalization technique.
\end{itemize}

The remainder of this paper is organized as follows: Section \ref{sec:related work} discusses relevant prior work. Sections \ref{sec:methodology} and \ref{sec:federated_learning} detail the datasets, preprocessing steps, model architecture, and the proposed FL framework including Random Subset Aggregation. Section \ref{sec:results} presents our experimental results, comparing localized, centralized, and various federated approaches. Section \ref{sec:discussion} interprets the findings and discusses implications. Finally, Section \ref{sec:conclusion} concludes the paper and suggests future directions.

\section{Related Work}
\label{sec:related work}

Several studies have used centralized, non-privacy preserving approaches for seizure prediction. One such study trained separate models on Siena and CHB-MIT, achieving strong within-dataset performance but poor generalization in a leave-one-patient-out setting, even after applying domain adaptation techniques \cite{jemal2024domain}. 
To address privacy concerns, another study explored FL on the CHB-MIT dataset \cite{aminifar2024privacy}. The authors treated each patient as a separate client, trained a collaborative model, and then applied local fine-tuning without data sharing.



While the previous study focused on patient-level federation within a single dataset, another work explored FL across multiple datasets \cite{MDPIs23146578}. Most similar to our setup, this study uses three public EEG datasets (CHB-MIT, Bonn, and NSC), treating each dataset as a single client, training locally on EEG segments, and averaging weights on a central server. The aggregated model's preictal scores are then combined with basic patient information for the final seizure-risk output. However, the paper does not specify which channels or window lengths were ultimately used. It also mixes label definitions, modeling preictal vs ictal for CHB-MIT but preictal vs interictal for Bonn and NSC. This lack of reproducibility makes direct comparison difficult and highlights the need for more detailed and transparent research in federated seizure prediction. 

Overall, there remains a clear lack of research in seizure prediction that explores collaboration across multiple hospitals, particularly in scenarios with significantly different EEG recordings between hospitals, leading to distributional differences between the clients. Our works aims to fill in this gap by simulating such a setting and showing how hospitals can train a robust model to perform seizure prediction in a privacy-preserving way, despite differing in patient characteristics. To the best of our knowledge, this is the first study to apply FL across all four datasets -- CHB-MIT, Helsinki, Siena, and NCH -- for seizure prediction.

\section{Seizure Prediction}
\label{sec:methodology}
\subsection{EEG Datasets and Preprocessing}

Table \ref{tab:dataset_summary} provides a summary of the four publicly available epileptic datasets used in our experiments:  CHB-MIT Scalp EEG \cite{guttag2010chb}, Helsinki University Hospital EEG \cite{stevenson2019dataset}, Siena Scalp EEG \cite{detti2020siena}, and Nationwide Children's Hospital (NCH) Sleep DataBank \cite{lee2022large}.  Only 14 patients in NCH who had some form of seizure and relevant EEG recordings were included in this study. For neonates, we report post-menstrual age in weeks~\cite{stevenson2019dataset}. The datasets vary in size and patient demographics, ranging from neonates to pediatrics to adults, making it an ideal setting to test for non-IID FL.

\begin{table}[htbp]
\caption{Summary of the epileptic EEG datasets. 
}
\centering
\setlength{\tabcolsep}{4pt}
\begin{tabular}{@{}llllp{2.2cm}@{}}
\toprule
\textbf{Dataset} & \textbf{Patients} & \textbf{Seizures}  & \textbf{Age Range} & \textbf{Note} \\
\midrule
CHB-MIT   & 22  & 182 & 1.5--22 yrs  & Anti-Seizure Medication Withdrawal \\
Helsinki  & 79 & 460  & 35--45 wks & NICU       \\
Siena     & 14    & 47& 20--71 yrs & -           \\
NCH       & 14  &  18& 2.5--23.3 yrs & Sleep study \\
\bottomrule
\end{tabular}
\label{tab:dataset_summary}
\end{table}

To keep our design as simple as possible, we focus on a single EEG channel (F3-C3) that is common among all our four datasets. For CHB-MIT, the F3-C3 bipolar channel is already available. For Siena and Helsinki, we construct this channel by computing the voltage difference between the individual electrodes F3 and C3. For NCH, we subtract the voltage of the C3-M2 channel from that of the F3-M2 channel.  We also apply a low-pass filter at 64 Hz before downsampling all signals to 128 Hz.

\subsection{Labeling and Segmenting EEG for Classification}
\begin{figure}[htbp]
    \centering
    \includegraphics[width=\linewidth]{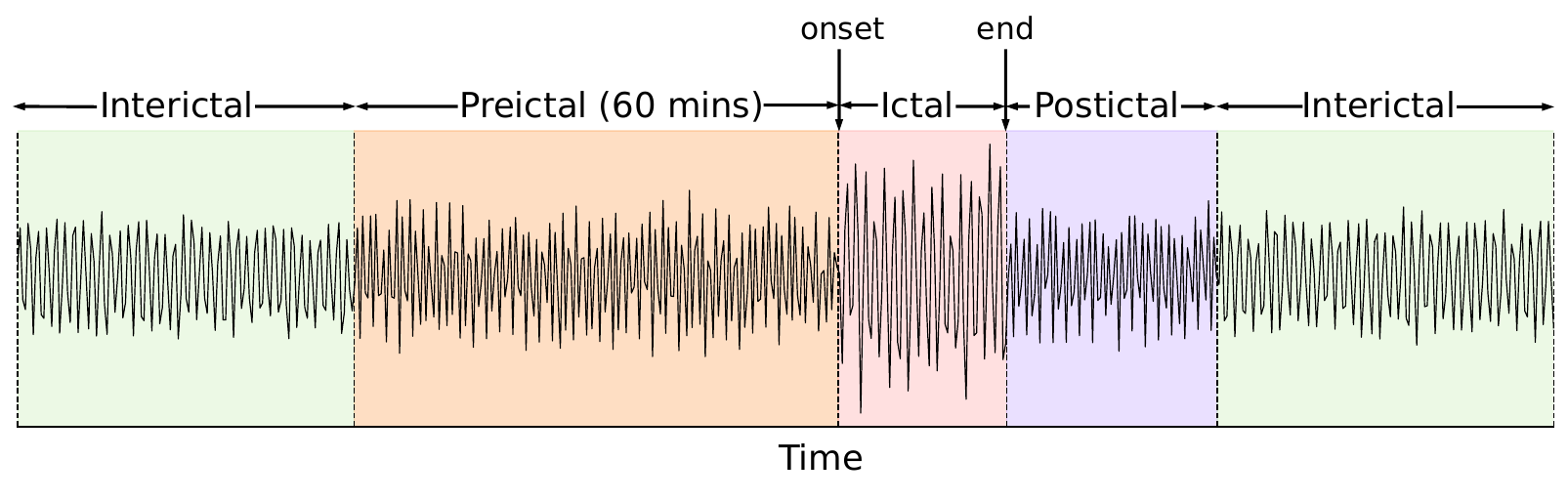} 
    \caption{Illustration of 4 epilepsy stages along with the seizure onset and end points. We perform binary classification between Preictal and Interictal stages of 2-second (sometimes overlapping) EEG segments.}
    \label{fig:seizure_stages}
\end{figure}

In this section, we outline our labeling policy for seizure prediction, focusing on four key stages of epilepsy: preictal, ictal, postictal, and interictal. Figure \ref{fig:seizure_stages} illustrates these stages.

The preictal window, the period before seizures that is used for anticipating seizures, has not been standardized and varies from minutes to hours \cite{preictalDesign}. Following the work of \cite{chbsiennaPreictal,pinto2022interpretable}, we set the preictal duration to be 1 hour.
We also discard the ictal state, which is when the seizure actually occurs. Our goal is to predict seizures early enough to warn patients in advance. Since the ictal phase means the seizure has already started, it's no longer useful for early intervention \cite{justifyNoDetection}. 

Similar to preictal, the choice of a postictal window -- the period after a seizure -- also varies across studies. 
However, while postictal states can last anywhere from minutes to days, it lasts between 5 and 30 minutes on average \cite{clinicalPostictalDefinition}. Based on this evidence, we define a postictal period of 10 minutes and exclude it from prediction task. 
This decision reflects the need to give the brain adequate time to return to baseline activity, reducing the likelihood of capturing abnormal postictal signals that could hinder model learning. 
Thus, we perform binary classification between  preictal and interictal, latter of which refers to the period of normal brain activity between seizures.

Finally, we segment the EEG signals into 2 seconds as in \cite{chbsiennaPreictal,2secondwindow}. 
Like in most seizure prediction studies, our preictal periods are shorter than interictal periods. To address this class imbalance, we apply data augmentation with a sliding window, which is commonly used in seizure prediction or detection \cite{chbsiennaPreictal,slidingWindowExample,helsinkiSeizure}. 
The stride was determined \textit{per patient} to roughly balance the final number of preictal and interictal examples, which is summarized by datasets in Table \ref{tab:label_distribution}. 

\begin{table}[ht]
\centering
\setlength{\tabcolsep}{4pt}
\caption{Number of examples after data augmentation.}
\begin{tabular}{@{}llll@{}}
\toprule
\textbf{Dataset} & \textbf{Window Overlap} & \textbf{Preictal Examples} & \textbf{Total Examples} \\
\midrule
CHB-MIT   & 60 - 95\% & 1,261,930 (45.76\%) & 2,757,911  \\
Helsinki  &   62\% &   101,335 (50.09\%) &   202,261  \\
Siena     &  7 - 91\%  &   183,492 (50.43\%) &   363,775 \\ 
NCH       &   75\% &   123,372 (44.71\%) &   275,922 \\
\bottomrule
\end{tabular}

\label{tab:label_distribution}
\end{table}

\subsection{Deep Learning Model Architecture}
We adopt an efficient and lightweight deep learning model called TinySleepNet \cite{TinySleepNet}, applying minor modifications to its architecture (e.g. kernel size, number of filters, removing dropout layers). TinySleepNet has been shown to generalize well across datasets with varying characteristics and recording setups for automatic sleep stage scoring based on raw single-channel EEG \cite{TinySleepNet}. Thus, with only 330,242 parameters and applicable on single-channel EEG, our model can demonstrate the potential of FL in environments with limited computational resources, such as clinics or edge devices.

\section{Federated Learning}
\label{sec:federated_learning}

We consider $K=4$ clients or hospitals (Table \ref{tab:dataset_summary}), each having access to only its own private EEG recordings. A central server (aggregator) facilitates model training without direct access to any raw EEG recordings. Let 
$\mathcal{D}_k = \{(\mathbf{x}_i^{(k)}, y_i^{(k)})\}_{i=1}^{n_k}$ be the dataset held by client $k$, where EEG segment $\mathbf{x} \in \mathbb{R}^{d}$ ($d=256$ for 2 seconds) and binary label $y \in \{0, 1\}$, where $y=0$ denotes an interictal segment and $y=1$ denotes a preictal segment. 

We follow the standard horizontal FL paradigm \cite{mcmahan2017communication}, where all clients share the same feature space (EEG channel) but hold different samples. A well-known challenge in this setup is the performance degradation under non-IID data distributions, which can cause significant model disparity across clients \cite{zhu2021federated, karimireddy2020scaffold}. In our case, EEG recordings differ significantly across hospitals, possibly due to variations in acquisition protocols, patient demographics, and clinical focus.

\subsection{Privacy-Preserving Global Normalization}
\label{sec:globalnorm}
Since EEG signal amplitudes may vary widely across datasets, we recommend normalization before training, either 1) a local standardization or 2) a local min-max normalization followed by a secure global standardization. Here, we describe one way to
compute the \textit{global} mean and standard deviation of the EEG signal across all clients, without revealing any individual hospital’s data. To achieve this, we adopt a commonly used secure aggregation approach based on zero-sum masking~\cite{bonawitz2016practical}.

Let $\mathbf{x}_i^{(k)} = [x_{i, j}^{(k)}]_{j=1}^{d}$, \(N = \sum_{k=1}^K  n_k\) and $N' = dN$. The goal is to securely compute:
\begin{itemize}
    \item the global mean: \(\mu = \frac{1}{N'} \sum_{k=1}^K \sum_{i=1}^{n_k} \sum_{j=1}^{d} x_{i,j}^{(k)}\),
    \item the global standard deviation:
    \[
        \sigma = \sqrt{\frac{1}{N'} \sum_{k=1}^K \sum_{i=1}^{n_k}\sum_{j=1}^{d}  \bigl(x_{i,j}^{(k)} - \mu\bigr)^2}
    \]
\end{itemize}
\subsubsection{Masking Mean Values} Each client \(k\) computes the sum of all their EEG vectors \(s_k = \sum_{i=1}^{n_k} \sum_{j=1}^{d} x_{i,j}^{(k)}\) and their sample count \(n_k\). They each generate two random masks \(\alpha_k,\delta_k \in \mathbb{R}\) such that \(\sum_{k=1}^K \alpha_k = 0\) and \(\sum_{k=1}^K \delta_k = 0\). Clients send masked values \(s_k + \alpha_k\), \(n_k + \delta_k\) to the server. Since masks cancel out, the server computes global mean as:
\begin{equation}
S = \sum_{k=1}^K s_k, \quad N' = \sum_{k=1}^K d \cdot n_k, \quad \mu = \frac{S}{N'}
\end{equation}
\subsubsection{Masking Variance Terms} Upon receiving \(\mu\) from the server, each client computes \(v_k = \sum_{i=1}^{n_k}\sum_{j=1}^{d} (x_{i,j}^{(k)} - \mu)^2\), generates a zero-sum mask \(\varepsilon_k \in \mathbb{R}\) such that \(\sum_{k=1}^K \varepsilon_k = 0\), and sends \(v_k + \varepsilon_k\) to the server. The server computes:
\begin{equation}
v = \frac{1}{N'} \sum_{k=1}^K v_k, \quad \sigma = \sqrt{v}
\end{equation}

\subsubsection{Local Normalization} The server broadcasts \(\mu\) and \(\sigma\) to all clients. Finally, each client locally normalizes their samples: 
\begin{equation}
x_{i,j}^{(k)} \leftarrow \frac{x_{i,j}^{(k)} - \mu}{\sigma}
\end{equation}

This approach assumes synchronous participation from all clients during the normalization phase with no dropouts. Each client follows a pairwise key agreement protocol (e.g., Diffie–Hellman) with every other client to derive shared random seeds. These seeds are expanded into mask vectors using a cryptographically secure pseudo-random generator (PRG). For each pair \((p, q)\), client \(p\) adds the PRG-generated vector \(\mathbf{r}_{p,q}\) while client \(q\) subtracts it, ensuring the masks cancel out when aggregated. This zero-sum mechanism, commonly used in privacy-sensitive domains such as healthcare \cite{antunes2022federated, liang2020we}, ensures that the server recovers only the global sums, while individual contributions remain hidden \cite{bonawitz2016practical}.

\subsection{Unweighted and Weighted FedAvg Algorithm}
We describe FedAvg \cite{mcmahan2017communication}, a widely adopted aggregation method in FL.

\subsubsection{Client Updates}

At communication round $t$, the server broadcasts the current global model parameters $\mathbf{w}^{(t)}$ to all clients. Each client $k$ then performs $E$ epochs of stochastic gradient descent on its local objective
\begin{equation}
F(\mathbf{w};~\mathcal{D}_k) = \frac{1}{n_k} \sum_{(\mathbf{x},y)\in \mathcal{D}_k} \ell\big(f(\mathbf{w}; \mathbf{x}),\, y\big),
\end{equation}
where $\ell(\cdot,\cdot)$ is the cross‐entropy loss and $f(\mathbf{w}; \mathbf{x})$ is our TinySleepNet model. Starting from $\mathbf{w}^{(t)}$, client $k$ computes
\begin{equation}
\mathbf{w}_k^{(t+1)} = \mathbf{w}^{(t)} - \eta \,\nabla F(\mathbf{w}^{(t)};~\mathcal{D}_k).
\end{equation}

\subsubsection{Aggregation}

Each client returns its updated weights $\mathbf{w}_k^{(t+1)}$ to the server. The server then computes the average 
\begin{equation}
\mathbf{w}^{(t+1)} \;=\; \frac{1}{K}\sum_{k=1}^K \,\mathbf{w}_k^{(t+1)}, \label{eq:fedavg_simple_avg}
\end{equation}
or the weighted average
\begin{equation}
\mathbf{w}^{(t+1)} \;=\; \sum_{k=1}^K \frac{n_k}{N}\,\mathbf{w}_k^{(t+1)}.\label{eq:fedavg_weighted_avg}
\end{equation}
 Weighting each client’s model update according to its local dataset size ensures that hospitals with larger datasets contribute proportionally more to the global model. However, in settings where dataset sizes vary significantly across clients, this strategy may lead to the under-representation of smaller clients, potentially biasing the global model toward the dominant data sources.

\subsubsection{Communication Rounds}

We run for $T$ communication rounds. In each round, all clients participate. Between rounds, only model weights are exchanged, no raw EEG or private patient data ever leave a hospital.

\subsection{Random Subset Aggregation}
\label{sec:undersampling}

Because each hospital’s EEG recording hardware and patient population differ, the local datasets $\{\mathcal{D}_k\}$ exhibit non‐IID characteristics. Even with global normalization, these differences, combined with substantial variation in dataset sizes (most notably, CHB-MIT having significantly more samples than the others), can lead to biased training in the federated setting, as demonstrated by our results in Section \ref{sec:results}. We provide a detailed analysis of client-wise performance variation and training behavior in the same section.

To mitigate the effects of data imbalance and distributional heterogeneity, we propose a \textit{Random Subset Aggregation} strategy to ensure that each hospital contributes equally in every local update. Our approach is similar to the sample-level balancing proposed in \cite{qi2023fedsampling}, which aims to achieve fairer and more stable federated training under data heterogeneity. However, while \cite{qi2023fedsampling} relies on probabilistic sampling following a global sample quota, we adopt a fixed-size sampling strategy for each client.

Let $M$ be a fixed subset size such that $M \le \min_k n_k$ (e.g., equal to the smallest hospital dataset). At each communication round $t$ and for each local epoch $e$, client $k$:

\begin{enumerate}
  \item Samples a subset without replacement 
\begin{equation} 
    \widetilde{\mathcal{D}}_k^{(t,e)} \subset \mathcal{D}_k
    \quad\text{of size}\quad
    \bigl|\widetilde{\mathcal{D}}_k^{(t,e)}\bigr| = M.
\end{equation}
  \item Performs a local update on this subset:
\begin{equation} 
    \mathbf{w}_k^{(t,e)} = \mathbf{w}_k^{(t,e-1)}
      - \eta\,\nabla F\bigl(\mathbf{w}_k^{(t,e-1)};\,\widetilde{\mathcal{D}}_k^{(t,e)}\bigr),
\end{equation}

  where
\begin{equation} 
    F(\mathbf{w};~\widetilde{\mathcal{D}}_k)
      = \frac{1}{M} \sum_{(\mathbf{x},y)\in \widetilde{\mathcal{D}}_k}
        \ell\bigl(f(\mathbf{w}; \mathbf{x}),y\bigr).
\end{equation}

\end{enumerate}

After $E$ epochs of subset-based updates, each client returns its final model parameters $\mathbf{w}_k^{(t,E)}$ to the server. Since all clients train on the same number of samples ($M$), the global model $\mathbf{w}^{(t+1)}$ is updated by taking a simple average across client parameters, as in Eq. \ref{eq:fedavg_simple_avg}.
This approach makes sure that, regardless of the original dataset sizes, each hospital has equal influence on the global model at every round. Algorithm \ref{alg:fedavg_undersample} outlines our full training loop with the undersampling approach.

\begin{algorithm}[t]
\caption{FedAvg with Random Subset Aggregation}
\label{alg:fedavg_undersample}
\begin{algorithmic}[1]
\State {\bf Input:} Initial parameters $\mathbf{w}^{(0)}$, epochs $E$, learning rate $\eta$, rounds $T$, subset size $M$
\For{communication round $t=0$ {\bf to} $T-1$}
  \State Server broadcasts $\mathbf{w}^{(t)}$ to all clients
  \For{each client $k$ {\bf in parallel}}
    \State $\mathbf{w}_k^{(t,0)} \gets \mathbf{w}^{(t)}$
    \For{local epoch $e=1$ {\bf to} $E$}
      \State Sample $\widetilde{\mathcal{D}}_k^{(t,e)} \subset \mathcal{D}_k$ where  $|\widetilde{\mathcal{D}}_k^{(t,e)}|=M$
      \State $\mathbf{w}_k^{(t,e)} \gets \mathbf{w}_k^{(t,e-1)} 
             - \eta\,\nabla F(\mathbf{w}_k^{(t,e-1)};\widetilde{\mathcal{D}}_k^{(t,e)})$
    \EndFor
    \State Client $k$ sends $\mathbf{w}_k^{(t,E)}$ to server
  \EndFor
  \State Server aggregates by averaging:
\begin{equation} 
    \mathbf{w}^{(t+1)} \gets \frac{1}{K}\sum_{k=1}^K \mathbf{w}_k^{(t,E)} \label{eq:fedavg_random_subset}
\end{equation}
\EndFor
\State {\bf Output:} $\mathbf{w}^{(T)}$
\end{algorithmic}
\end{algorithm}

While a natural choice for $M$ is the size of the smallest dataset, which ensures that all hospitals can sample without replacement, this approach may inadvertently reveal the dataset size of the smallest hospital, leading to potential side-information leakage. To avoid this, we treat $M$ as a tunable parameter. In Section~\ref{sec:results}, we provide an empirical analysis using different values of $M$, demonstrating its impact on the performance of the global model.

\section{Results}
\label{sec:results}
 All results are reported over 5 runs of 80-10-10 train-validation-test splits, with mean (standard deviation) in \%. Our code is available at \url{https://github.com/sauravpandey123/SeizureFed}.

\subsection{Localized Learning}
\label{subsec:localized_learning}

\begin{figure}[htbp]
    \centering
    \includegraphics[width=0.9\linewidth]{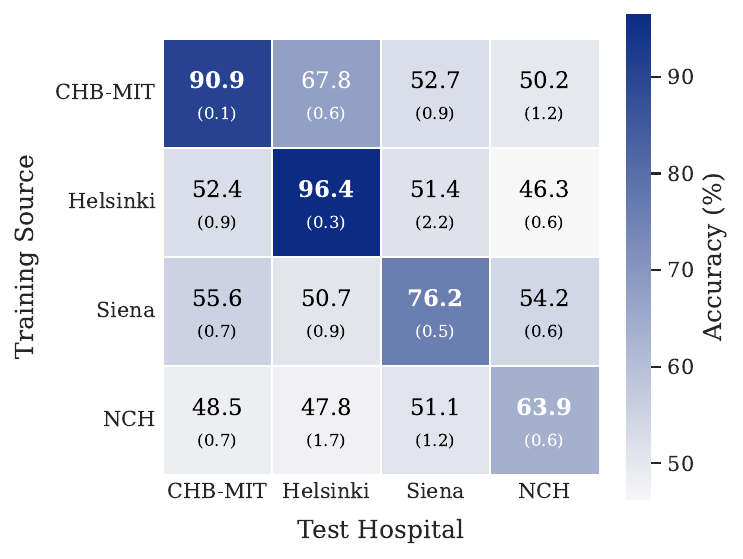} 
    \caption{Cross-hospital generalization performance of local models. Each row corresponds to a model trained using data from a single hospital, and each column represents the test performance on another hospital’s test data. Values indicate test accuracy (in \%), with diagonal entries representing local model performance.}
    \label{fig:heatmap_cross_hospital}
\end{figure}


We first trained a local model on each hospital's train data and tested it on all four test sets to evaluate cross-dataset performance.
From Figure \ref{fig:heatmap_cross_hospital}, we observe that each model performs quite well on its local data but the accuracy drops significantly when tested on other sites. For instance, the Helsinki-trained model performs poorly on NCH (46.3\%) and CHB-MIT (52.4\%), despite an impressive 96.4\% accuracy on its own test set. Similar trends are seen across all other clients too, which suggests that models trained on a single hospital struggle to generalize across different populations or recording setups.

These results highlight the effects of data heterogeneity, e.g., patient demographics, hardware differences. Even though we applied preprocessing to eliminate basic voltage range mismatches, it cannot solve distribution shifts, which continue to limit generalization. This motivates the need for more collaborative learning techniques, such as FL, where we can train a single global model that is both robust and generalizable across all clients.

\subsection{Centralized Learning}

Next, we trained a single, centralized model on the union of the four training sets. This setting serves as an ideal baseline in which data can be freely shared without privacy constraints, offering an upper bound on performance achievable by privacy-sensitive models. 
As shown in Table \ref{tab:centralized_performance}, the centralized model achieves strong overall performance across all test sets, as measured by accuracy, AUROC, and F1-score. These results show the benefit when full data is available, as the model is able to learn from a more balanced and diverse representation of patient populations. However, in practical settings, each hospital does not want to share its patients' private data with other hospitals due to privacy, legal, and ethical concerns. 
This limitation motivates the need for privacy-preserving approaches such as FL, whose implementation we discuss in Section \ref{subsec:federated_learning_results}. 

\begin{table}[t]
\centering
\caption{Performance of the centralized learning model on test sets.}
\label{tab:centralized_performance}
\begin{tabular}{lccc}
\toprule
\textbf{Test Set} & \textbf{Accuracy (\%)} & \textbf{F1-Score (\%)} & \textbf{AUROC (\%)} \\
\midrule
Pooled   & 81.2 (0.3) & 79.9 (0.2) & 90.4 (0.2) \\
\midrule
CHB-MIT      & 82.7 (0.4) & 81.2 (0.3) & 91.6 (0.2) \\
Helsinki & 91.9 (0.2) & 91.7 (0.2) & 96.6 (0.1) \\
Siena    & 75.8 (0.2) & 76.2 (0.1) & 85.1 (0.2) \\
NCH      & 66.4 (0.2) & 64.5 (0.3) & 73.4 (0.3) \\
\midrule
\textbf{Macro Avg.} & 79.6 (0.3) & 78.7 (0.2) & 87.4 (0.2) \\
\bottomrule
\end{tabular}
\end{table}

We use the results from the centralized approach (\textbf{Central}) as an ideal upper bound for our FL experiments. 
By comparing FL results against Table \ref{tab:centralized_performance}, we gain valuable insights into how well FL performs in comparison to a fully shared data setting. 

\subsection{Federated Learning}
\label{subsec:federated_learning_results}

We now evaluate three variants of FedAvg: \textbf{Unweighted average}~\eqref{eq:fedavg_simple_avg}, \textbf{Weighted average}~\eqref{eq:fedavg_weighted_avg}, and our proposed \textbf{Random subset aggregation} (Algorithm~\ref{alg:fedavg_undersample}). 

Before fixing our final setup, we also experimented with varying the number of local epochs \(E\) per round—both globally and even custom \(E_k\) per client, to see if more local passes could boost convergence or personalization.  None of these variants yielded any consistent improvement, and in some cases longer local training actually hurt underrepresented sites. Accordingly, all of the federated results use \(E=1\), which we found to be the best after extensive experiments.

\subsubsection{FedAvg with Unweighted and Weighted Average}

\begin{table}[htbp]
\centering
\caption{Performance of FedAvg (unweighted average) on test sets}
\label{tab:baseline_simple_avg}
\begin{tabular}{lccc}
\toprule
\textbf{Test Set} & \textbf{Accuracy (\%)} & \textbf{F1-Score (\%)} & \textbf{AUROC (\%)} \\
\midrule
Pooled   & 77.5 (0.6) & 75.3 (0.8)  & 86.3 (0.8) \\
\midrule
CHB-MIT  & 82.6 (0.7) & 80.9 (0.8)  & 91.0 (0.6) \\
Helsinki & 53.8 (1.9) & 45.4 (20.0) & 59.2 (4.2) \\
Siena    & 68.5 (0.5) & 68.0 (1.0)  & 75.4 (0.8) \\
NCH      & 55.4 (0.7) & 42.2 (11.7) & 57.3 (1.5) \\
\midrule
\textbf{Macro Avg.} & 65.1 (11.6) & 59.1 (16.0) & 70.7 (13.7) \\
\bottomrule
\end{tabular}
\end{table}

\begin{table}[htbp]
\centering
\caption{Performance of FedAvg (weighted average) on test sets}
\label{tab:baseline_fedavg}
\begin{tabular}{lcccc}
\toprule
\textbf{Test Set} & \textbf{Accuracy (\%)} & \textbf{F1-Score (\%)} & \textbf{AUROC (\%)} \\
\midrule
Pooled   & 80.8 (0.2) & 79.2 (0.3) & 89.0 (0.7) \\
\midrule
CHB-MIT  & 89.0 (0.1) & 87.9 (0.0)  & 95.2 (0.1) \\
Helsinki & 50.8 (2.0) & 33.8 (13.7) & 51.8 (3.6) \\
Siena    & 58.9 (1.7) & 63.1 (0.5)  & 63.5 (1.3) \\
NCH      & 50.6 (2.1) & 43.1 (7.8)  & 49.5 (1.7) \\

\midrule
\textbf{Macro Avg.} & 62.3 (15.7) & 57.0 (20.7) & 65.0 (18.2) \\
\bottomrule
\end{tabular}
\end{table}

\begin{figure*}[ht]
    \centering
    \includegraphics[width=\textwidth]{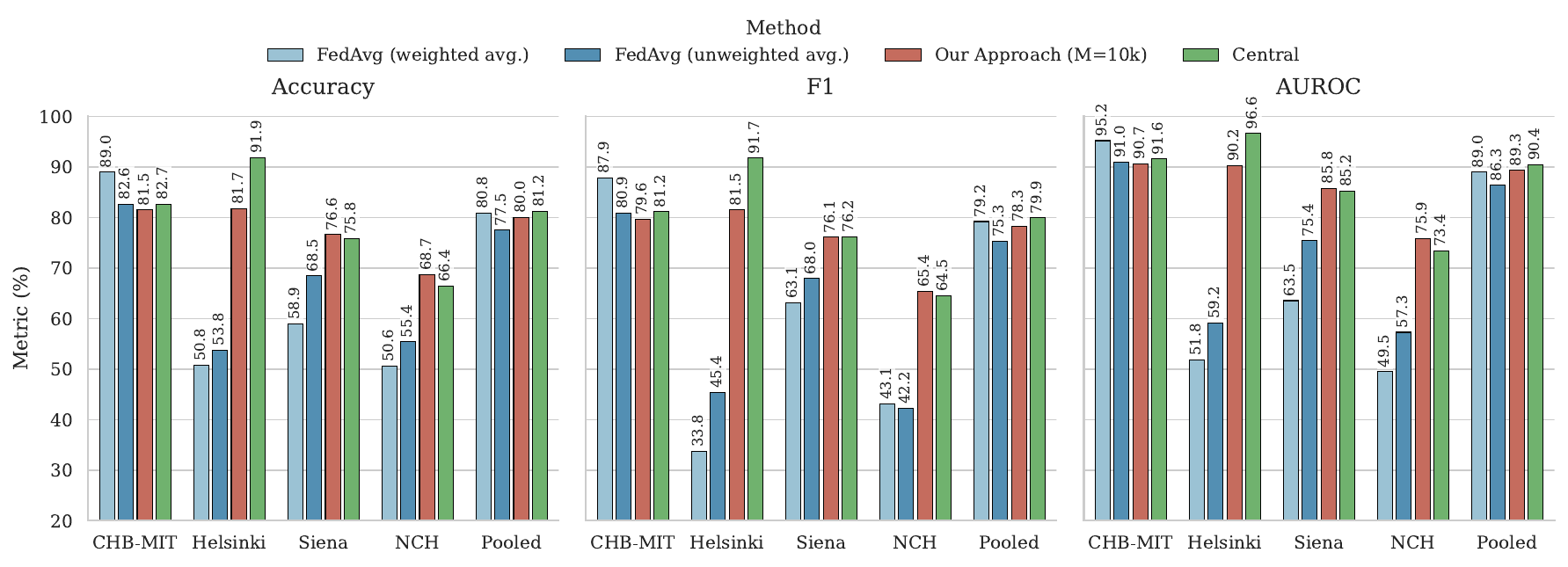} 
    \caption{Classification performance of the best performing subset size ($M=10k$) for the proposed approach compared to the FedAvg with and without sample size weighting, as well as the non-private central model. Metrics are evaluated on each hospital's test data and on the pooled test set. Results are the mean over multiple runs, with bars annotated with mean values (in \%).}
    \label{fig:10k_vs_baselines}
\end{figure*}

Table~\ref{tab:baseline_simple_avg} shows the performance of averaging on the pooled test set and on each hospital individually. Although the pooled accuracy reaches 77.5\%, the macro‑averaged accuracy across hospitals is only 65.1\%, due to low scores on the under represented sites Helsinki (53.8\%) and NCH (55.4\%). Similarly, Table~\ref{tab:baseline_fedavg} shows that using a weighted average improves pooled accuracy to 80.8\%, but the macro average performance drops even further to 62.3\%. Most importantly, the model behaves similarly to a completely random classifier on smaller sites such as Helsinki (50.8\%) and NCH (50.6\%). These results confirm that, without incorporating additional measures for balancing, both standard federated training approaches are dominated by the largest dataset (CHB‑MIT) and fail to properly represent smaller clients.

\subsubsection{Random Subset Aggregation (Proposed)}

Figure~\ref{fig:10k_vs_baselines} compares the classification performance of the proposed method and several baseline approaches: FedAvg (with and without data-size weighting), our Random Subset Aggregation with \(M=10,000\), and the centralized non‑private model. Results are reported for each hospital individually as well as on the pooled test set.

On the pooled test set, all methods yield relatively similar performance. Weighted FedAvg achieves 80.8\% accuracy, 79.2\% F1, and an AUROC of 89.0\%. Simple averaging with FedAvg appears to be the worst-performing method on the pooled set, reaching 77.5\% accuracy, 75.3\% F1-score, and an AUROC of 86.3\%. The centralized model performs slightly better, with 81.2\% accuracy, 79.9\% F1, and an AUROC of 90.4\%. Our proposed method yields 80.0\% accuracy, 78.3\% F1-score, and 89.3\% AUROC -- offering performance that remains competitive with the centralized ideal.

On the under-represented sites, FedAvg (both weighted and unweighted) performs poorly across all metrics, achieving only 50.8–53.8\% accuracy, 33.8–45.4\% F1, and 51.8–59.2\% AUROC on Helsinki, and 50.6–55.4\% accuracy, 42.2–43.1\% F1, and 49.5–57.3\% AUROC on NCH. In contrast, our method yields substantial improvements -- achieving 81.7\% accuracy, 81.5\% F1, and 90.2\% AUROC on Helsinki. While this still falls short of the centralized upper bound, it shows a significant gain over all federated baselines. On NCH, our method achieves 68.7\% accuracy, 65.4\% F1, and 75.9\% AUROC, outperforming all baselines and even surpassing the accuracy of the locally trained model (63.9\%) and the centralized model (66.4\%). These results demonstrate that our Random Subset Aggregation strategy is effective in mitigating non-IID imbalance, and not only substantially improves global generalization, but also enables underrepresented clients, in some cases, to outperform both local and centralized baselines.

Interestingly, even on Siena -- the second most well-represented site -- our method outperforms all baselines, achieving 76.6\% accuracy, 76.1\% F1, and 85.8\% AUROC. It even slightly exceeds the centralized model, which yields 75.8\% accuracy, with a marginally higher F1-score of 76.2 and an AUROC of 85.2, and also slightly outperforms the localized model's accuracy of 76.2\%. In comparison, weighted FedAvg reaches only 58.9\% accuracy, 63.1\% F1, and 63.5\% AUROC, while unweighted FedAvg performs moderately better with 68.5\% accuracy, 68.0\% F1, and 75.4\% AUROC.

On the most well-represented site, CHB‑MIT, the weighted FedAvg method performs best, achieving 89.0\% accuracy, 87.9\% F1‑score, and 95.2\% AUROC. Both the unweighted FedAvg and the centralized model yield comparable results, with the former reaching 82.6\% accuracy, 80.9\% F1, and 91.0\% AUROC, and the latter attaining 82.7\% accuracy, 81.2\% F1, and 91.6\% AUROC. Our proposed method, while slightly underperforming on CHB‑MIT with 81.5\% accuracy, 79.6\% F1, and 90.7\% AUROC, still closely matches the centralized and unweighted FedAvg models.

Overall, the proposed method provides a more balanced global model -- offering substantial improvements on the under-represented clients (Helsinki and NCH), and even enhancing performance on a well-represented site like Siena, with only a modest trade-off in performance on CHB‑MIT. These results highlight the effectiveness of our method in mitigating data imbalance and handling cross-site heterogeneity in federated seizure prediction.

\subsection{Effect of $M$ in Random Subset Aggregation}

\begin{figure*}[htbp]
    \centering
    \includegraphics[width=\textwidth]{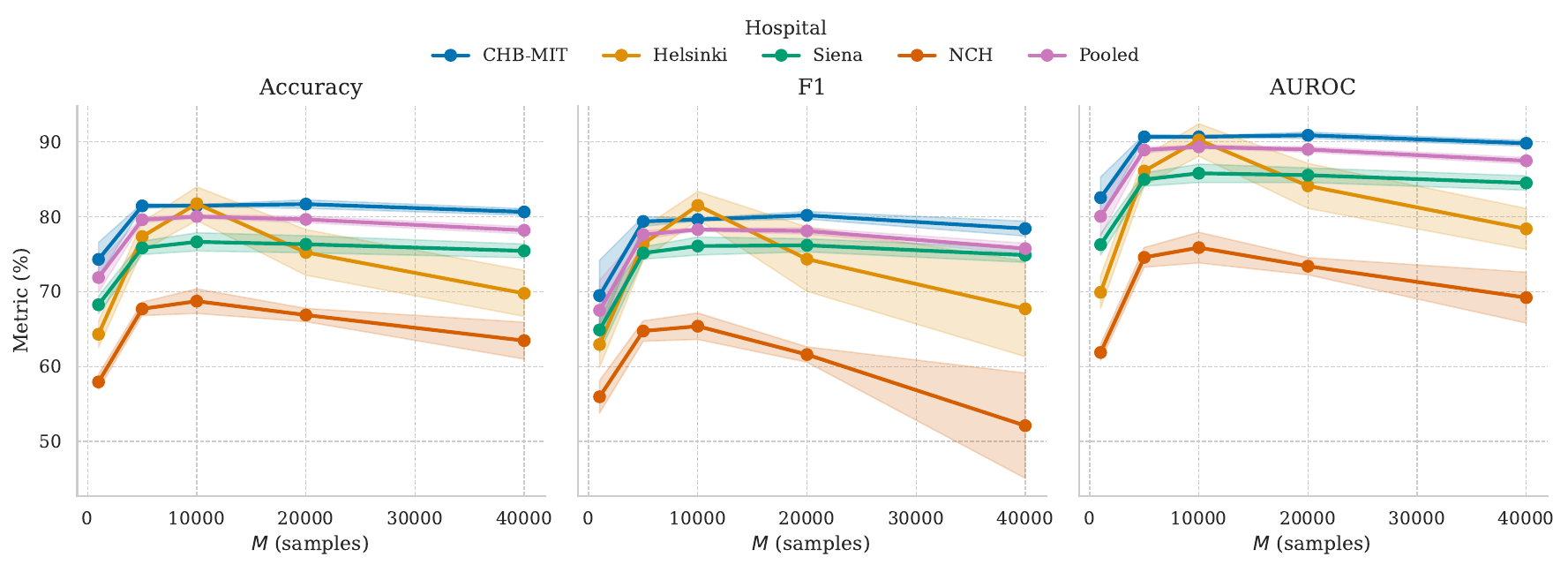} 
    \caption{Classification performance of the global models trained using the Random Subset Aggregation method across hospitals for varying subset sizes. Metrics are evaluated on each hospital's test data and on the pooled test set. The shaded area represents the standard deviation, with results shown as mean.}
    \label{fig:undersampling_for_m}
\end{figure*}

Figure~\ref{fig:undersampling_for_m} and Table~\ref{tab:macro_avg_performance} shows how the accuracy of the global model varies across hospitals and on the pooled test set as we change the subset size \(M\). At very small \(M\) (1,000), the global model remains heavily biased toward CHB‑MIT, achieving high accuracy there but still underperforming on Helsinki and NCH. As \(M\) increases, the accuracy curves for the under‑represented sites rise sharply. For all \(M\) values, performance on the well-represented sites remains stable around the 76\% range for Siena and around 82\% range for CHB-MIT. We believe that this divergence in sensitivity to \(M\) reflects the degree of initial under‑representation. Helsinki, which started with the lowest accuracy under both weighted (50.8\%) and unweighted (53.8\%) FedAvg (see Tables~\ref{tab:baseline_simple_avg} and~\ref{tab:baseline_fedavg}), benefits most from subset aggregation, reaching up to 81.5\% accuracy at $M = 10{,}000$.

\begin{table}[ht]
\centering
\caption{Macro-averaged performance across CHB-MIT, Helsinki, NCH, and Siena for different subset sizes $M$}
\label{tab:macro_avg_performance}
\begin{tabular}{rcccc}
\toprule
\textbf{$M$} & \textbf{Accuracy (\%)} & \textbf{F1-Score (\%)} & \textbf{AUROC (\%)} \\
\midrule
1,000            & 66.2 (6.0) & 63.3 (4.9) & 72.6 (7.7) \\
5,000            & 75.6 (5.0) & 73.9 (5.5) & 84.1 (5.9) \\
\textbf{10,000}  & 77.1 (5.3) & 75.6 (6.2) & 85.6 (6.0) \\
20,000           & 75.0 (5.3) & 73.1 (7.0) & 83.5 (6.3) \\
40,000           & 72.3 (6.4) & 68.3 (10.1) & 80.5 (7.7) \\
\bottomrule
\end{tabular}
\end{table}

Table~\ref{tab:macro_avg_performance} presents these trends into macro‑averaged metrics across all four hospitals. Macro accuracy steadily climbs from 66.2\% at \(M=1,000\) to a peak of 77.1\% at \(M=10,000\), then plateaus and begins to drop as \(M\) grows further. A similar pattern holds for F1‑score (from 63.3\% to 75.6\%), and AUROC (from 72.6\% to 85.6\%). These gains show that while equalizing client contributions (as done across all $M$ values) is important for fairness, selecting an appropriate subset size is equally important for achieving a generalizable and robust model. At very small or very large $M$, even with balanced updates, the model still underperforms on Helsinki and NCH, whereas intermediate values like $M = 10{,}000$ provide a better balance between fairness and generalizability.

We also observe that the standard deviation in F1-score on NCH and Helsinki, increases at higher values of $M$, particularly beyond $M = 10{,}000$. However, similar variance is also present in both weighted and unweighted FedAvg baselines (see Tables~\ref{tab:baseline_simple_avg} and~\ref{tab:baseline_fedavg}), suggesting that this instability is not specific to our aggregation strategy. Instead, it likely reflects the underlying heterogeneity and limited sample sizes in these datasets, which can lead to higher variability in classification performance across runs.

Together, the Fig.~\ref{fig:undersampling_for_m} and Table~\ref{tab:macro_avg_performance} confirm that Random Subset Aggregation at an intermediate \(M\) effectively mitigates the skew introduced by non‑IID, imbalanced data, yielding strong and generalizable performance across diverse hospital settings without excessively penalizing the largest dataset. For completeness, Table~\ref{tab:results_vs_m} reports the exact accuracy, F1-score, and AUROC values for each subset size \(M\) on all individual hospital test sets and the pooled test set.

\begin{table}[t]
\centering
\caption{Per‐hospital and pooled test performance for varying subset sizes \(M\).}
\begin{tabular}{l r c c c}
\toprule
\textbf{Test Set} & \textbf{$M$} & \textbf{Accuracy (\%)} & \textbf{F1-Score (\%)} & \textbf{AUROC (\%)} \\
\midrule
\multirow{5}{*}{Pooled}
    & 1,000  & 71.9 (2.0) & 67.5 (4.0) & 80.0 (2.6) \\
    & 5,000  & 79.6 (0.4) & 77.6 (0.5) & 88.9 (0.4) \\
    & 10,000 & 80.0 (0.2) & 78.3 (0.2) & 89.3 (0.2) \\
    & 20,000 & 79.6 (0.4) & 78.1 (0.4) & 89.0 (0.4) \\
    & 40,000 & 78.2 (0.5) & 75.7 (0.7) & 87.5 (0.4) \\
\midrule
\multirow{5}{*}{CHB‑MIT}
    & 1,000  & 74.3 (2.3) & 69.5 (4.7) & 82.5 (2.8) \\
    & 5,000  & 81.4 (0.3) & 79.4 (0.6) & 90.7 (0.3) \\
    & 10,000 & 81.5 (0.2) & 79.6 (0.2) & 90.7 (0.1) \\
    & 20,000 & 81.7 (0.6) & 80.2 (0.5) & 90.9 (0.4) \\
    & 40,000 & 80.6 (0.4) & 78.4 (1.0) & 89.8 (0.4) \\
\midrule
\multirow{5}{*}{Helsinki}
    & 1,000  & 64.3 (1.8) & 62.9 (3.1) & 69.9 (2.2) \\
    & 5,000  & 77.3 (1.9) & 76.4 (2.2) & 86.1 (1.9) \\
    & 10,000 & 81.7 (2.3) & 81.5 (1.9) & 90.2 (2.2) \\
    & 20,000 & 75.2 (3.0) & 74.3 (4.3) & 84.1 (3.0) \\
    & 40,000 & 69.8 (3.1) & 67.7 (6.4) & 78.4 (2.7) \\
\midrule
\multirow{5}{*}{Siena}
    & 1,000  & 68.2 (1.2) & 64.9 (3.0) & 76.3 (1.4) \\
    & 5,000  & 75.8 (0.9) & 75.2 (0.8) & 85.0 (0.9) \\
    & 10,000 & 76.6 (1.2) & 76.1 (1.2) & 85.8 (1.2) \\
    & 20,000 & 76.3 (1.1) & 76.2 (0.9) & 85.5 (1.0) \\
    & 40,000 & 75.4 (0.9) & 74.9 (1.0) & 84.5 (1.0) \\
\midrule
\multirow{5}{*}{NCH}
    & 1,000  & 57.9 (1.0) & 56.0 (2.2) & 61.9 (1.1) \\
    & 5,000  & 67.7 (0.9) & 64.7 (1.4) & 74.6 (1.3) \\
    & 10,000 & 68.7 (1.6) & 65.4 (1.8) & 75.9 (2.0) \\
    & 20,000 & 66.9 (0.9) & 61.6 (1.0) & 73.4 (1.2) \\
    & 40,000 & 63.5 (2.5) & 52.1 (7.0) & 69.2 (3.4) \\
\bottomrule
\end{tabular}
\label{tab:results_vs_m}
\end{table}

\section{Discussion}
\label{sec:discussion}

Our experiments show that, when substantial distributional shifts exist across hospitals, baseline FL approaches -- including FedAvg (with and without data-size weighting) and the centralized non‑private model -- perform worse than a hospital’s own locally trained model on its native test data. This indicates that each hospital still achieves its highest performance by training solely on its own EEG recordings, despite participating in cross‑site collaboration. At first glance, this might suggest that collaboration and FL is unnecessary, since each site’s hardware setup and patient demographics will most likely remain stable over time, and locally trained models remain optimal for on‑site deployment.

However, our results also show that these same local models fail to generalize when applied to data from other hospitals. A hospital aiming to deploy a seizure prediction system in a new clinical setting (potentially one with different acquisition protocols or patient populations) would find that neither its own model nor any single site model performs adequately. This lack of cross‑site generalizability highlights the value of a collaborative approach. By training a shared global model, hospitals can obtain a more robust predictor that transfers more effectively to unseen environments. 

In this context, our balanced FL strategy based on Random Subset Aggregation proves to be a promising technique for handling non‑IID settings. It produces a global model whose performance across all hospitals approaches that of the centralized ideal, while preserving each hospital’s data privacy. Furthermore, our global model sometimes even outperforms a client's locally trained model, which highlights that under appropriate aggregation, FL can yield benefits beyond what local data alone can offer. Overall, our global model offers a single, privacy‑preserving solution that generalizes well, unlike any of the individual hospital models or standard FL approaches like FedAvg.

Looking forward, these findings motivate further research into personalized federated algorithms that can adapt the shared global model back to each hospital’s local distribution, as well as techniques to mitigate cross‑site distribution shifts (e.g., domain adaptation). Such extensions could help close the remaining gap between cross‑site and local performance, yielding privacy‑preserving models that are simultaneously optimal for both native and unseen deployment settings.

\section{Conclusion}
\label{sec:conclusion}
We demonstrated the challenges and opportunities of applying FL to seizure prediction across heterogeneous EEG datasets. Our findings reveal that standard FL methods, such as FedAvg, are insufficient in highly non-IID settings, leading to models that disproportionately favor dominant datasets. The proposed Random Subset Aggregation technique achieved significantly improved generalization across diverse clinical populations and recording conditions. These results highlight the importance of designing balanced FL strategies to enable the development of privacy-preserving models in realistic multi-hospital environments. 

\bibliographystyle{ieeetr}
\bibliography{references}

\end{document}